%% file: root.tex
\documentclass[letterpaper, 10 pt, conference]{ieeeconf}

\usepackage{amsmath}
\usepackage{amssymb}
\usepackage{graphicx}
\usepackage{xcolor}
\usepackage{booktabs}
\usepackage{tabularx}
\usepackage{caption}
\usepackage{array}
\usepackage{makecell}
\usepackage{multirow}
\usepackage{float}
\usepackage{cuted}
\usepackage{url}

\RequirePackage{etoolbox}
\AtEndPreamble{
    \usepackage[capitalize]{cleveref}
    \crefname{section}{Sec.}{Secs.}
    \Crefname{section}{Section}{Sections}
    \Crefname{table}{Table}{Tables}
    \crefname{table}{Tab.}{Tabs.}
}

\title{\LARGE \bf Seeing Space and Motion: Enhancing Latent Actions with Geometric and Dynamic Awareness for Vision-Language-Action Models}

\author{
    Zhejia Cai$^{1,2,*}$, 
    Yandan Yang$^{1}$, 
    Xinyuan Chang$^{1}$, 
    Shiyi Liang$^{1,3,*}$, \\
    Ronghan Chen$^{1}$, 
    Feng Xiong$^{1,\ddagger}$, 
    Mu Xu$^{1}$, 
    Ruqi Huang$^{2,\dagger}$
}

\begin{document}

\maketitle

\begin{strip}
    \centering
    \includegraphics[width=0.98\linewidth]{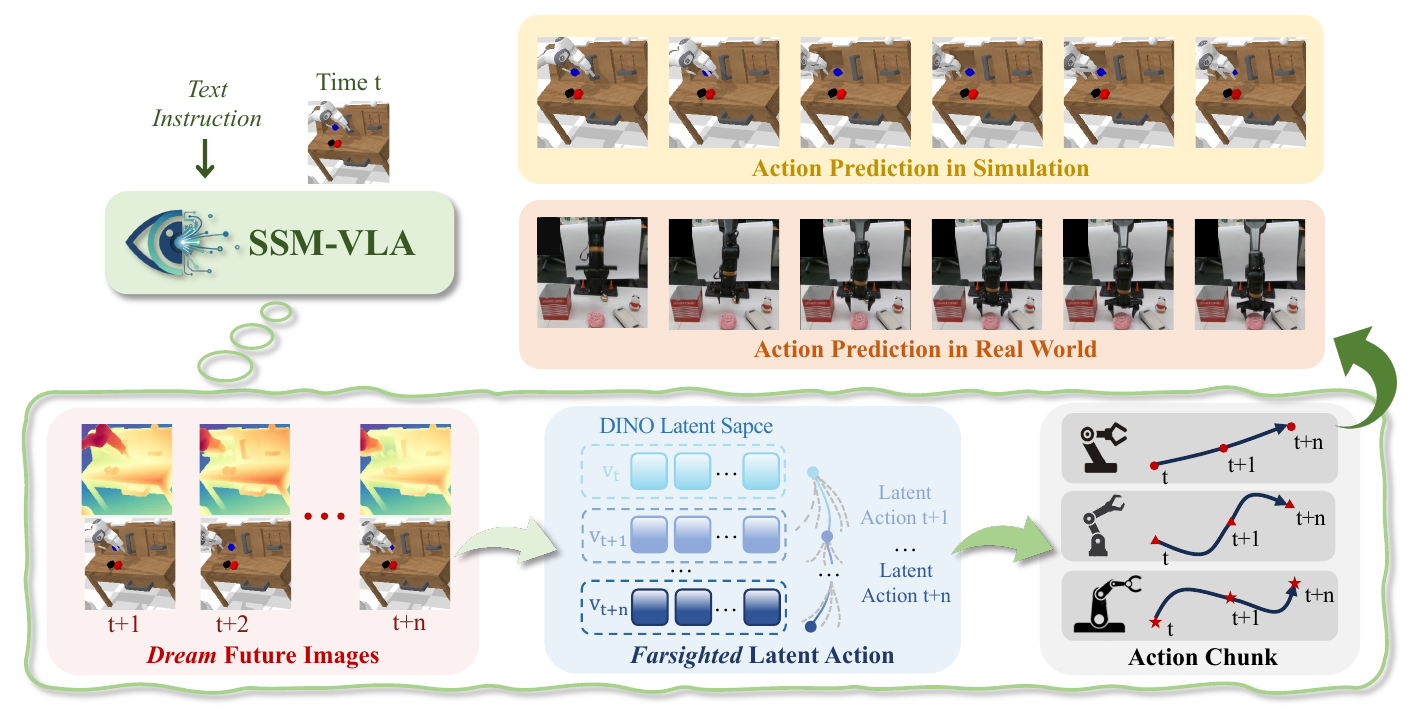}
    \captionof{figure}{
        \textbf{Illustration of the end-to-end causal reasoning pipeline within a single SSM-VLA model}, encompassing three core stages: 1) Future observation prediction, generating a visual chain-of-thought to enable interpretable and temporally coherent reasoning; 2) Farsighted latent action modeling, integrating spatial and temporal dynamics for effective long-horizon policy planning; 3) Modular action chunk prediction, supporting cross-platform generalization across diverse robotic embodiments. 
        Experiments on both real-world and simulated environments demonstrate SSM-VLA’s robustness and practical effectiveness.}
    \label{fig:overview}
\end{strip}

\thispagestyle{empty}
\pagestyle{empty}
\footnotetext[1]{AMAP, Alibaba Group.}
\footnotetext[2]{Tsinghua Shenzhen International Graduate School, Tsinghua University.}
\footnotetext[3]{School of Software Engineering, Xi'an Jiaotong University.}
\def\thefootnote{*}\footnotetext{This work was conducted during the internship at Alibaba Group.}
\def\thefootnote{\textdagger}\footnotetext{Corresponding author: \url{ruqihuang@sz.tsinghua.edu.cn}}
\def\thefootnote{\textdaggerdbl}\footnotetext{Project leader.}
\def\thefootnote{\arabic{footnote}}

\input{sec/0_abstract}  
\input{sec/1_intro}
\input{sec/2_related_work}
\input{sec/3_method}
\input{sec/4_evaluation}

\input{sec/5_conclusion}

\addtolength{\textheight}{0cm}
\bibliographystyle{IEEEtran}
\bibliography{references}

\end{document}

%% file: sec/0_abstract.tex
\begin{abstract}
Latent Action Models (LAMs) enable Vision-Language-Action (VLA) systems to learn semantic action representations from large-scale unannotated data. 
Yet, we identify two bottlenecks of LAMs: 1) the commonly adopted end-to-end trained image encoder suffers from poor spatial understanding; 2) LAMs can be fragile when input frames are temporally distant, leading to limited temporal perception. 
Such factors inevitably hinder stable and clear action modeling.
To this end, we propose Farsighted-LAM, a latent action framework with geometry-aware spatial encoding and multi-scale temporal modeling, capturing structural priors and dynamic motion patterns from consecutive frames. 
We further propose SSM-VLA, an end-to-end VLA framework built upon Farsighted-LAM, which integrates structured perception with a visual Chain-of-Thought module to explicitly reason about environmental dynamics, enhancing decision consistency and interpretability. 
We validate SSM-VLA on multiple VLA tasks in both simulation and real-world settings, and achieve state-of-the-art performance. 
Our results demonstrate that our strategy of combining geometry-aware modeling, temporal coherence, and explicit reasoning is effective in enhancing the robustness and generalizability of embodied intelligence.
\end{abstract}

%% file: sec/1_intro.tex
%%%%%%%%%%%%%%%%%%%%%%%%%%%%%%%%%%%%%%%%%%%%%%%%%%%%%%%%%%%%%%%%%%%%%%%%%%%%%%%%
\section{Introduction}

Latent Action Models (LAMs) have emerged as a promising paradigm for Vision-Language-Action (VLA) systems, enabling self-supervised learning of compact, semantic action representations from large-scale, unannotated vision-language data. 
By capturing spatial configurations and motion patterns, LAMs facilitate action consequence reasoning and future state anticipation without requiring explicit embodiment or fine-grained action labels. 
Such properties allow robots to acquire generalizable policies from internet-scale data with minimal real-world interaction. 
This shift toward scalable, data-driven learning paves the way for more adaptable and generalist agents.

However, existing LAMs remain limited in robust embodied reasoning due to two critical shortcomings: 1) Inadequate spatial understanding, where direct RGB encoding biases latent actions toward surface textures, neglecting geometric structure such as object relations and scene layout; and 2) Limited temporal perception, as most methods rely on sparse, two-frame inputs (e.g., UniVLA~\cite{bu2025learning}, Moto-GPT~\cite{chen2024moto}), failing to capture both long-term dynamics and fine-grained motion transitions. 
This dual deficiency leads to unstable and semantically ambiguous action representations, undermining decision reliability.

To address the above issues, we propose Farsighted-LAM, a latent action modeling framework that enhances spatial and temporal fidelity through two key designs: 1) Geometry-aware spatial encoding using DINOv2~\cite{oquab2023dinov2} features, which encode structural priors—e.g., spatial layouts, implicit depth, and object relations—enabling geometrically consistent and semantics-rich scene understanding; and 2) Multi-scale temporal modeling via consecutive frame sequences, capturing both sustained motion trends and transient interactions (e.g., contacts, manipulations), thereby improving temporal coherence and prediction stability. 
Together, these advances enable more structured and dynamic environment modeling.

Building upon Farsighted-LAM, we further introduce \textbf{S}eeing \textbf{S}pace and \textbf{M}otion (\textbf{SSM})\textbf{-VLA}, an end-to-end VLA framework that integrates structured perception with a Chain-of-Thought (CoT) reasoning module to explicitly simulate environmental dynamics before action execution, enhancing interpretability and physical plausibility. 
We validate SSM-VLA in both simulation and real-world robotic tasks, achieving state-of-the-art performance on CALVIN ABC-D benchmark. 
Our results demonstrate that our strategy of combining geometry-aware modeling, temporal coherence, and explicit reasoning is effective in enhancing the robustness and generalizability of embodied intelligence.

In summary, our contributions are as follows:
\begin{itemize}
\item We propose Farsighted-LAM, a latent action model with enhanced spatial understanding and multi-scale temporal modeling, enabling robust representation of scene structure and dynamic motion patterns.
\item We propose SSM-VLA, an end-to-end VLA framework that integrates Farsighted-LAM for geometry-aware spatiotemporal modeling with a visual CoT module, enhancing decision consistency and interpretability.
\item Through comparisons with competitive baseline models, we show that SSM-VLA achieves state-of-the-art performance on a challenging VLA benchmark.
\end{itemize}

%% file: sec/2_related_work.tex
\section{Related Works}

\subsection{Vision-Language-Action Models}
A dominant paradigm in robot learning trains end-to-end policies that directly map high-dimensional sensory inputs to low-level actions. Pioneered by generalist agents like Gato~\cite{reedgeneralist} and established in robotics by Octo~\cite{ghosh2024octo} and RT-1~\cite{brohan2022rt}, this approach has been scaled via fine-tuning vision-language models (e.g., RT-2~\cite{zitkovich2023rt}, OpenVLA~\cite{kim2025openvla}) and generative methods such as Diffusion Policy~\cite{chi2023diffusion} and its transformer-based variants~\cite{roboflamingo,hou2025dita, ke20253d}. The strength of this model lies in integrating perception, reasoning, and control into a single framework. However, this direct end-to-end prediction method still faces three fundamental challenges: insufficient support from observational information for action decision-making, over-coupling of policy learning with physical carrier characteristics, and difficulty in effectively utilizing unlabeled video data rich in physical and interaction dynamics.

\subsection{Latent Action Pretraining}
To bridge the gap between representation learning and control, prior work pre-trains visual representations from video (e.g., Genie~\cite{bruce2024genie}, Dynamo~\cite{cui2024dynamo}, R3M~\cite{nair2023r3m}) for downstream adaptation. Building on this, Latent Action Models (LAMs) learn robot-independent latent actions from observation pairs through an inverse dynamics model, captures the intent of state transitions, and uses a lightweight network to decode them into embodied motor commands.
LAMs have evolved from early frameworks like IGOR~\cite{chen2024igor} and LAPO~\cite{chen2022lapo} to large-scale unsupervised pre-training with vision-language models such as LAPA~\cite{yelatent}, reducing reliance on labeled data. Extensions like UniAct~\cite{zheng2025universal} and UniVLA~\cite{bu2025learning} enable cross-embodiment generalization via universal or task-centric latent actions, achieving state-of-the-art efficiency.
VideoWorld~\cite{Ren_2025_CVPR} relies on multi-frame images to predict latent dynamics but focuses on knowledge acquisition instead of CoT reasoning in Vision-Language-Action models. Our method, Farsighted-LAM, addresses this issue and significantly enhances the model's geometric cognition and dynamic awareness.

\subsection{Visual Expectation Enhancement}
Existing work, known as Inverse Dynamics Models (IDM), focuses on using visual expectations to enhance the context of the current observation for action prediction. VideoAgent~\cite{soni2024videoagent} is a self-improving system that refines generated video plans for robot control by leveraging environmental feedback to correct hallucinations and boost task success. Gen2Act~\cite{bharadhwaj2024gen2act} enables robots to generalize to novel tasks by conditioning a single policy on generated human videos. Seer~\cite{tianpredictive} introduces an end-to-end paradigm that learns scalable robot policies by predicting actions from its own forecasted visual states. The Video Prediction Policy (VPP)~\cite{hu2024video} achieves state-of-the-art robotic manipulation by conditioning its actions on the rich, "predictive visual representations" extracted from pre-trained Video Diffusion Models. Our method further enhances action guidance by forecasting future visual states imbued with geometric priors, leading to more precise control in 3D space.

%% file: sec/3_method.tex
\section{Method}

In this section, we introduce the details of our proposed SSM-VLA. First, we design Farsighted Latent Action Model to learn latent action with enhancement of dynamic spatial information in \ref{subsec:farsighted_lam}. 
Then we introduce the overall VLA policy in \ref{subsec:cascaded_vla} with three stages of prediction: VisualCoT, latent action, and action prediction.

\subsection{Farsighted Latent Action Model}
\label{subsec:farsighted_lam}

\begin{figure}[t]
    \centering
    \includegraphics[width=0.8\columnwidth]{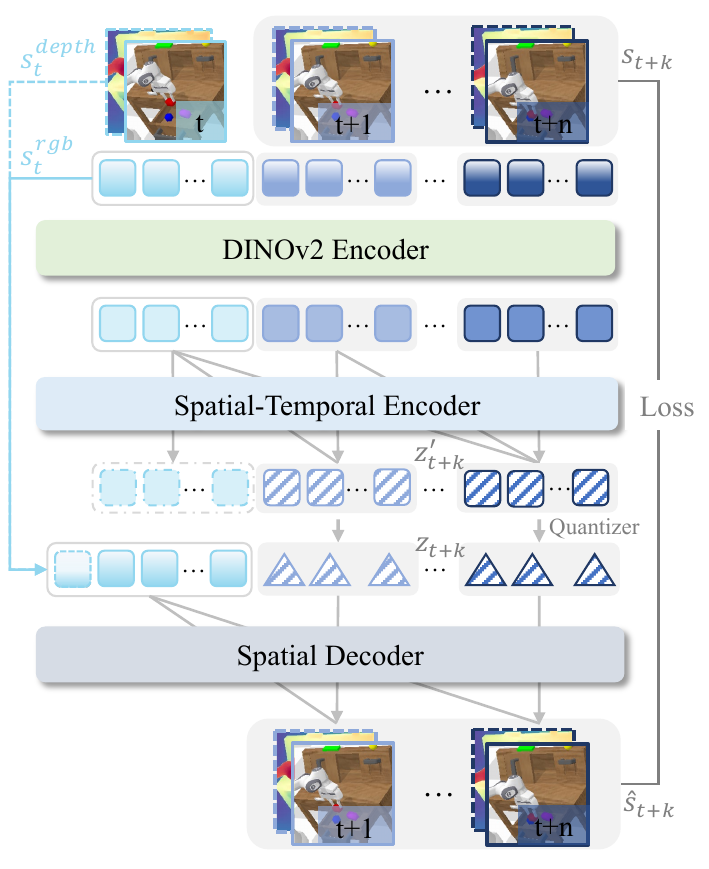}
    \caption{\textbf{Architecture of our Farsighted Latent Action Model.} The encoder takes DINOv2 features of the current frame $s_t$ and multiple future keyframes to predict a sequence of latent actions. The decoder then uses the current frame $s_t$ and a quantized latent action $z_{t+k}$ to reconstruct the corresponding future frame $\hat{s}_{t+k}$.}
    \label{fig:latent_action_model}
\end{figure}

Latent action model(LAM) learns a structured latent action space from unlabeled videos, aiming to improve the generalization ability of VLA systems. 
It usually takes in a pair of observations, marked as $s_t,s_{t+K}$, to compute the latent action $a_t$ at time $t$. Here $K$ is a fixed time interval between these two observations. 
Here we modify this process and design this model to enhance both spatial reasoning and dynamic modeling ability for the latent action, as illustrated in~\cref{fig:latent_action_model}.

\subsubsection{\textbf{Encoder}}

In contrast to conventional LAMs operating on a pair of observations, our model extends the receptive field of LAM and simultaneously processing a sequence of future $N$ key frames $\{s_{t+i}\}_{i=1}^N$ to predict a corresponding sequence of latent actions in a single forward pass. This helps the model capture both detailed and long-horizon motion information of the dynamic scene.
Meanwhile, to reason with deeper awareness of the structural priors in spatial domain, we take not only RGB image but also depth for the observation $s_t$, which is denoted as $s_t = (s_t^{rgb},s_t^{depth})$. RGB image is used as input and depth is utilized for additional supervision.
We further leverage features $v_{t}$ from a frozen DINOv2 encoder, denoted as $\Phi_V$ to ground the representation in a geometrically and semantically rich space. 
Specifically, we first extract the corresponding visual feature for each RGB frame, where $v_t = \Phi_V(s_t^{rgb})$, and get a sequence of feature $\{v_{t+i}\}_{i=0}^N$.

These features, along with a set of $N$ learnable \textit{latent action queries}, $\mathcal{Q} = \{q_k\}_{k=1}^N$, are processed by a spatio-temporal transformer $\Psi_{\text{ST}}$ to jointly encode the space and motion. For each query $q_k \in \mathcal{Q}$, the transformer generates a sequence of continuous latent vectors, each representing a continuous latent action for the future timestep $t+k$:

\begin{equation}
    z'_{t+k} = \Psi_{\text{ST}}(\{v_{t+i}\}_{i=0}^k, q_k),
\end{equation}
where $k \in \{1,2,...,N\}$.
Subsequently, each continuous latent vector $z'_{t+k}$ is quantized to a discrete token $z_{t+k}$ via a nearest-neighbor lookup in a learned codebook $\mathcal{C} \subset \mathbb{R}^D$:
\begin{equation}
    z_{t+k} = \underset{c \in \mathcal{C}}{\arg\min} \| z'_{t+k} - c \|_2
\end{equation}
The final encoded feature is a sequence of discrete tokens $\{z_{t+k} \}_{k=1}^N$, which constitutes the quantized representation of the future action plan and serves as the latent action.

\subsubsection{\textbf{Decoder}}

The decoder is responsible for validating the semantic and dynamic content of a learned latent action by translating it back into the visual domain. Here we implement the decoder as a spatial transformer, denoted as $\Psi_{\text{S}}$. It predicts the future observation $\hat{s}_{t+k}$ at time $t+k$ given only the initial observation $s_t$ and a discrete latent action $z_{t+k}$ of the future time step $t+k$. We emphasize that the predicted observation $\hat{s}_{t+k}$ includes both RGB $\hat{s}_{t+k}^{rgb}$ and depth $\hat{s}_{t+k}^{depth}$. This makes sure that the latent action has learned not only the dynamic information of visual texture but also the spatial structure of the scene. The generation process is formulated as:
\begin{equation}
    \hat{s}_{t+k} = (\hat{s}_{t+k}^{rgb}, \hat{s}_{t+k}^{depth}) = \Psi_{\text{S}}(s_t^{rgb},s_t^{depth}, z_{t+k})
\end{equation}
Different from the encoder, here we restrict the input of the decoder to the tuple of $(s_t^{rgb}, z_{t+k})$, making the decoder blind to the ground-truth target observation $s_{t+k}$ or any intermediate observations $\{s_{t+j}\}_{j=1}^{k-1}$. This constraint prevents the decoder from learning shortcut mappings with adjacent frames, which conversely enforces the encoder to embed more space and motion information into the latent action $z_{t+k}$. We count on the latent action $z_{t+k}$ to bridge the gap from $s_t$ to $s_{t+k}$ in the decoding process as well as the overall VLA policy in ~\cref{subsec:cascaded_vla}.

\subsubsection{\textbf{Reconstruction Loss}}
\label{subsec:recon_loss}
We propose a multi-modal reconstruction loss $\mathcal{L}_{\text{rec}}$ to supervise the farsighted latent action model. Since the decoder generates both RGB $\hat{s}_{t+k}^{rgb}$ and depth $\hat{s}_{t+k}^{depth}$, we conduct loss on these two modal predictions according to the ground-truth observations $s_{t+k}^{rgb}$ and $s_{t+k}^{depth}$. 

The photometric loss $\mathcal{L}_{\text{rgb}}$ combines $L2$ loss with the LPIPS perceptual loss~\cite{zhang2018perceptual} with weight $\lambda_{LPIPS}$:
\begin{equation}
    \mathcal{L}_{\text{rgb}}(s, \hat{s}) = \|\hat{s}^{\text{rgb}} - s^{\text{rgb}}\|_2^2 + \lambda_{\text{lpips}} \cdot \mathcal{L}_{\text{lpips}}(\hat{s}^{\text{rgb}}, s^{\text{rgb}})
\end{equation}
This constraint primarily ensures that the rendered output is photorealistic and captures the correct appearance, which is crucial for strengthening the model's understanding of semantic content like textures and object identities.

Then the depth loss $\mathcal{L}_{\text{depth}}$ is a gradient-aware logarithmic loss~\cite{turkulainen2025dn} that inversely weights the loss by the RGB image gradient:
\begin{equation}
\begin{split}
    \mathcal{L}_{\text{depth}}(s, \hat{s}) = \exp(-\|\nabla s^{\text{rgb}}\|) \\ 
    \cdot  \frac{1}{|\mathcal{P}|} \sum_{p \in \mathcal{P}} \log(1 + |\hat{s}^{\text{depth}}(p) - s^{\text{depth}}(p)|),
\end{split}
\end{equation}
where $p \in \mathcal{P}$ means each pixel of the depth map. This depth constraint enforces geometric consistency, which is fundamental to a robust understanding of the underlying 3D spatial structure.

The final reconstruction loss is a weighted sum over all $N$ future key frames. By integrating the photometric loss $\mathcal{L}_{\text{rgb}}$ and the depth loss $\mathcal{L}_{\text{depth}}$, we ensure that the model learns a representation that is faithful in both appearance and geometry. The hyperparameter $\lambda_{\text{d}}$ balances the relative importance of these two constraints:
\begin{equation}
    \mathcal{L}_{\text{rec}} = \sum_{k=1}^{N} \left( \mathcal{L}_{\text{rgb}}(s_{t+k}, \hat{s}_{t+k}) + \lambda_{\text{d}} \mathcal{L}_{\text{depth}}(s_{t+k}, \hat{s}_{t+k}) \right).
    \label{eq:recon_loss}
\end{equation}

\subsection{The Overall VLA Policy}
\label{subsec:cascaded_vla}

\begin{figure*}[t!]
    \centering
    \includegraphics[width=0.85\linewidth]{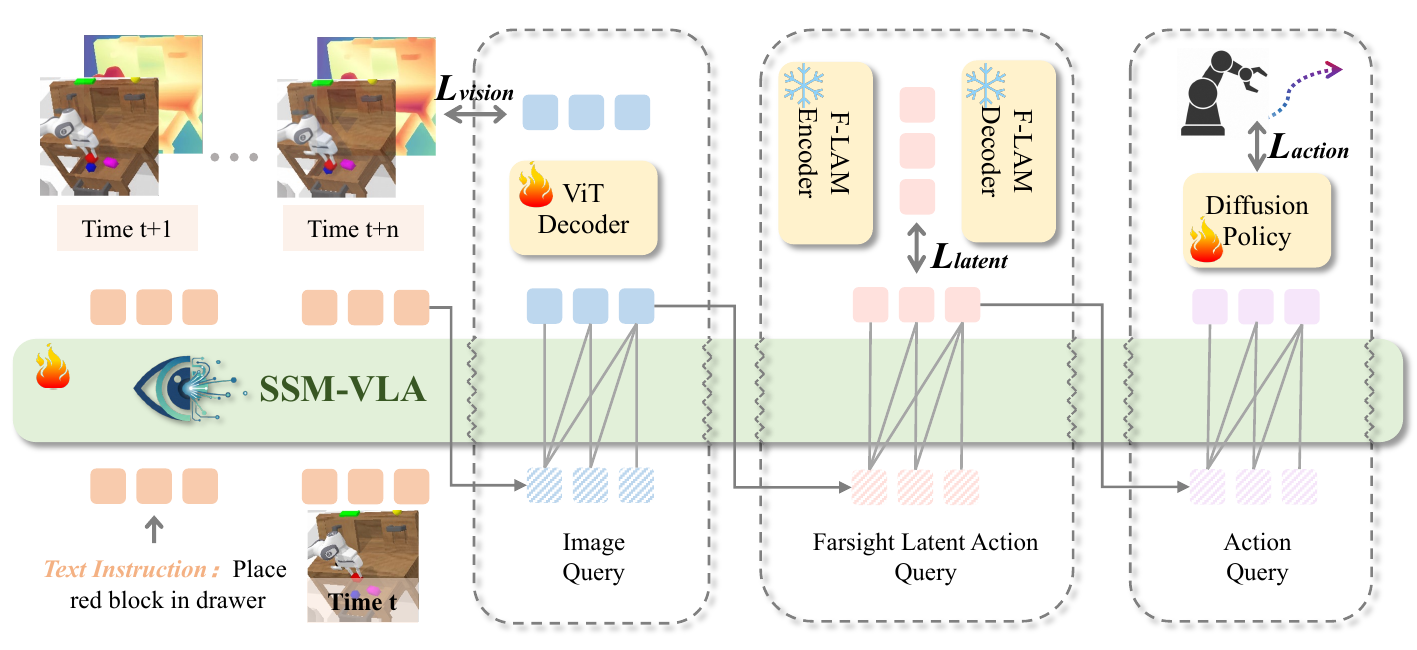}
    \caption{\textbf{The Three-Stage Cascaded VLA Policy.} Stage 1 predicts the immediate future observation $\hat{s}_{t+k}$. Stage 2 infers a long-horizon latent action plan $\{\hat{z}_{t+k}\}_{k=1}^{N}$. Stage 3 fuses all information to produce the final executable action $a_t$.}
    \label{fig:vla_policy}
\end{figure*}

As shown in ~\cref{fig:vla_policy}, the SSM-VLA model takes the current visual observation and natural language commands as input. Inspired by Moto-GPT~\cite{chen2024moto}, the model first predicts foresightful implicit actions as intermediate representations using a Foresight Implicit Action Model (F-LAM). These implicit actions remain semantically abstract and task-generalizable, effectively decoupling high-level task intent from low-level execution details, thereby reducing the policy's dependence on specific hardware configurations and significantly improving cross-platform transferability. Subsequently, an action query mechanism combined with a diffusion-based policy adapts the implicit actions to the target robot's action space.

Furthermore, we observe a significant modal gap in the direct mapping from “image \& instruction” to the final motor action, which can easily lead to optimization instability and decreased generalization ability. Inspired by FSDrive~\cite{zeng2025FSDrive}, we introduce a Visual CoT mechanism, using the prediction of future visual states (e.g., RGB or depth frames) as an intermediate inference step. This "imagine first, then act" paradigm both strengthens the model's spatiotemporal understanding and improves the accuracy and temporal coherence of generated actions significantly.

In summary, our model operates in three cascaded stages:
VisualCoT Prediction, Farsighted Latent Action Inference, and Action Generation. During fine-tuning, each stage takes a query vector as additional input and is supervised by a specific objective.
We detail the implementation of each stage here.

\subsubsection{\textbf{Stage 1: VisualCoT Prediction}}\par
A visual prediction module, $\mathcal{M}_{\text{vision}}$, takes the historical observations $s_{t-H:t}$ ($H$ is the history length) and language instruction $l$ to generate the next visual state:
\begin{equation}
    \hat{s}_{t+1} = \mathcal{M}_{\text{vision}}(s_{t-H:t}, l)
\end{equation}
This prediction is supervised by a vision loss, $\mathcal{L}_{\text{vision}}$. We adopt the same loss formulation as for reconstruction (see ~\cref{eq:recon_loss}), supervising both the predicted RGB image and depth map. This enables the VLA model to concurrently forecast both future semantic observations (via RGB) and geometric structure (via depth). For data with sensor depth, it is formally identical to the reconstruction loss $\mathcal{L}_{\text{rec}}$ applied to the single next frame:
\begin{equation}
    \mathcal{L}_{\text{vision}} = \mathcal{L}_{\text{rec}}(s_{t+1}, \hat{s}_{t+1})
    \label{eq:loss_vis}
\end{equation}
For data without sensor depth, since depth maps predicted from a single image by methods like DepthAnything~\cite{depth_anything_v2} are inherently normalized and lack a metric scale, it is necessary to align these varied predictions into a consistent world coordinate system. To this end, we follow \cite{turkulainen2025dn} and generate a pseudo-target. An initial estimate $D_{\text{mono}}$ is aligned to a sparse map $D_{\text{sparse}}$, obtained via SfM \cite{schoenberger2016sfm} using different views and poses of camera. The alignment is solved via closed-form linear regression over the set of sparse pixels $\mathcal{P}_{\text{sparse}}$:
\begin{equation}
    \hat{a}, \hat{b} = \arg \min_{a,b} \sum_{p \in \mathcal{P}_{\text{sparse}}} \|(a \cdot D_{\text{mono}}(p) + b) - D_{\text{sparse}}(p)\|_2^2
\end{equation}
Then we get the pseudo target of depth with $s^{depth} = \hat{a} \cdot D_{\text{mono}} + \hat{b}$ and apply the visual loss in~\cref{eq:loss_vis} on it

\subsubsection{\textbf{ Stage 2: Farsighted Latent Action Inference}}
\label{subsec:latent_from_VLA}

The latent prediction module, $\mathcal{M}_{\text{latent}}$, takes the historical context and the predicted next frame's features to infer a sequence of future action-intent distributions with length $N$:
\begin{equation}
    \hat{z}_{t+k} = \mathcal{M}_{\text{latent}}(s_{t-H:t}, l, \hat{s}_{t+1}, \{\hat{z}_{t+j}\}_{j=1}^{k-1}),
\end{equation}
where $k \in \{1,2,...,N\}$.
It is supervised by ground-truth latent action $z_{t+k}$, which is generated by the fixed encoder of the previously trained Farsighted Latent Action Model (F-LAM). F-LAM here takes in the ground-truth video frames. Note here the predicted $\hat{z}_{t+k} $ has been projected to the space of discrete latent action $z_{t+k}$. Thus, here we use a Cross-Entropy loss for the latent action:
\begin{equation}
    \mathcal{L}_{\text{latent}} = - \sum_{k=1}^{N} z_{t+k} \log(\hat{z}_{t+k})
\end{equation}

\subsubsection{\textbf{ Stage 3: Action Generation}}

 Then the action module, $\mathcal{M}_{\text{action}}$, takes a comprehensive context vector, including the historical context and the predicted latent action to generate the intermediate feature $c_t$ of the robot action:
\begin{equation}
    c_t = \mathcal{M}_{\text{action}}(s_{t-H:t}, l, \hat{s}_{t+1}, \{\hat{z}_{t+j}\}_{j=1}^{N})
\end{equation}
Then we utilize this feature $c_t$ as the condition of a conditional Flow Matching model $V_\theta$~\cite{lipmanflow} to predict the real action. The corresponding loss is:
\begin{equation}
    \mathcal{L}_{\text{action}} = \mathbb{E}_{\tau, \epsilon, a_t} \left [ \| V_\theta(\tau a_t + (1-\tau)\epsilon, \tau, c_t) - (\epsilon - a_t) \|_2^2 \right]
\end{equation}
where $V_\theta$ is a DiT network and $\epsilon \sim \mathcal{N}(0, \mathbf{I})$.

The entire VLA policy is fine-tuned by minimizing a composite loss, $\mathcal{L}_{\text{VLA}}$, which integrates the objectives from each training stage through a weighted summation:
\begin{equation}
    \mathcal{L}_{\text{VLA}} = \mathcal{L}_{\text{action}} + \lambda_{\text{latent}} \mathcal{L}_{\text{latent}} + \lambda_{\text{vision}} \mathcal{L}_{\text{vision}}
\end{equation}

\subsection{Multi-modal Synergistic Attention}
The cascaded architecture of SSM-VLA is implemented within a single, unified transformer through a carefully designed Multi-modal Synergistic Attention mechanism. Initially, historical visual tokens $(s_{t-H:t})$ and language tokens $(l)$ form a bi-directionally attentive core context, grounding the instruction in visual history. Subsequently, the visual prediction stage queries attend only to this core context to generate $\hat{s}_{t+1}$, ensuring the prediction is strictly a function of the past. The latent planning stage then takes a step further, with its queries attending to both the core context and the predicted frame $\hat{s}_{t+1}$ to produce the plan $\{\hat{z}_{t+j}\}_{j=1}^{N}$; a causal mask within these queries ensures the plan's temporal coherence. Finally, the action query $a_t$ acts as the final information aggregator, focusing on the core context, the predicted next frame, and the complete implicit plan. This progressively structured attention mechanism enables each component to specialize based on its preceding output, synergistically driving the reasoning capabilities of SSM-VLA.

%% file: sec/4_evaluation.tex
\section{Evaluations}
\input{Tab/calvin_main.tex}
\subsection{Implementation Details}
All models are implemented in PyTorch. Both models employ a cosine learning rate schedule with a 5\% linear warm-up phase. For the Latent Action Model, we use an AdamW optimizer~\cite{loshchilovdecoupled} with an initial learning rate of $10^{-4}$ and a weight decay of $10^{-5}$. The batch size is set to 256 while total training step is set to 100, and the size of codebook $\mathcal{C}$ is set to 32. For the CALVIN benchmark, this model is trained to encode the latent action between the current frame and the subsequent 3 frames, while for real-world data, it models the action between the current frame and the next 2 frames. Each frame is represented by 4 discrete tokens, and the flow matching head uses 10 denoising steps. The loss weights for this model are set as $\lambda_{\text{lpips}}=1$, $\lambda_{rgb}=1$, and $\lambda_{d}=0.01$. For the VLA Model, we also use an AdamW optimizer, but with an initial learning rate of $10^{-3}$ and a weight decay of $10^{-4}$. The batch size for this model is 64 while total training step is set to 30. Its loss weights are set as $\lambda_{\text{vision}}=0.1$ and $\lambda_{\text{latent}}=0.01$.
\subsection{Simulation Benchmark Experiments}
\subsubsection{Experiments Setup}
We evaluate our approach on the CALVIN~\cite{mees2022calvin} benchmark, which consists of 34 distinct manipulation tasks defined by open-ended language instructions, ranging from simple pick-and-place to complex articulated object manipulation. The benchmark utilizes a Franka Panda robotic arm across four tabletop environments. For our experiments, we trained policies on demonstration data from environments A, B, and C, and then undergo zero-shot evaluation in the unseen environment D. We utilized no-language-instruction data to pretrain our model following methods in Seer~\cite{tianpredictive}. The policy's generalization is rigorously tested on 1,000 unique instruction chains, each requiring the completion of five consecutive tasks. We utilized all of three types of signals (static camera, gripper camera and proprioceptive state) as input to achieve best performance. As presented in Table 1, the results on this demanding CALVIN ABC-D benchmark show that SSM-VLA achieves top performance. 
\subsubsection{Results}
\begin{figure}[t]
    \centering
    \includegraphics[width=0.95\columnwidth]{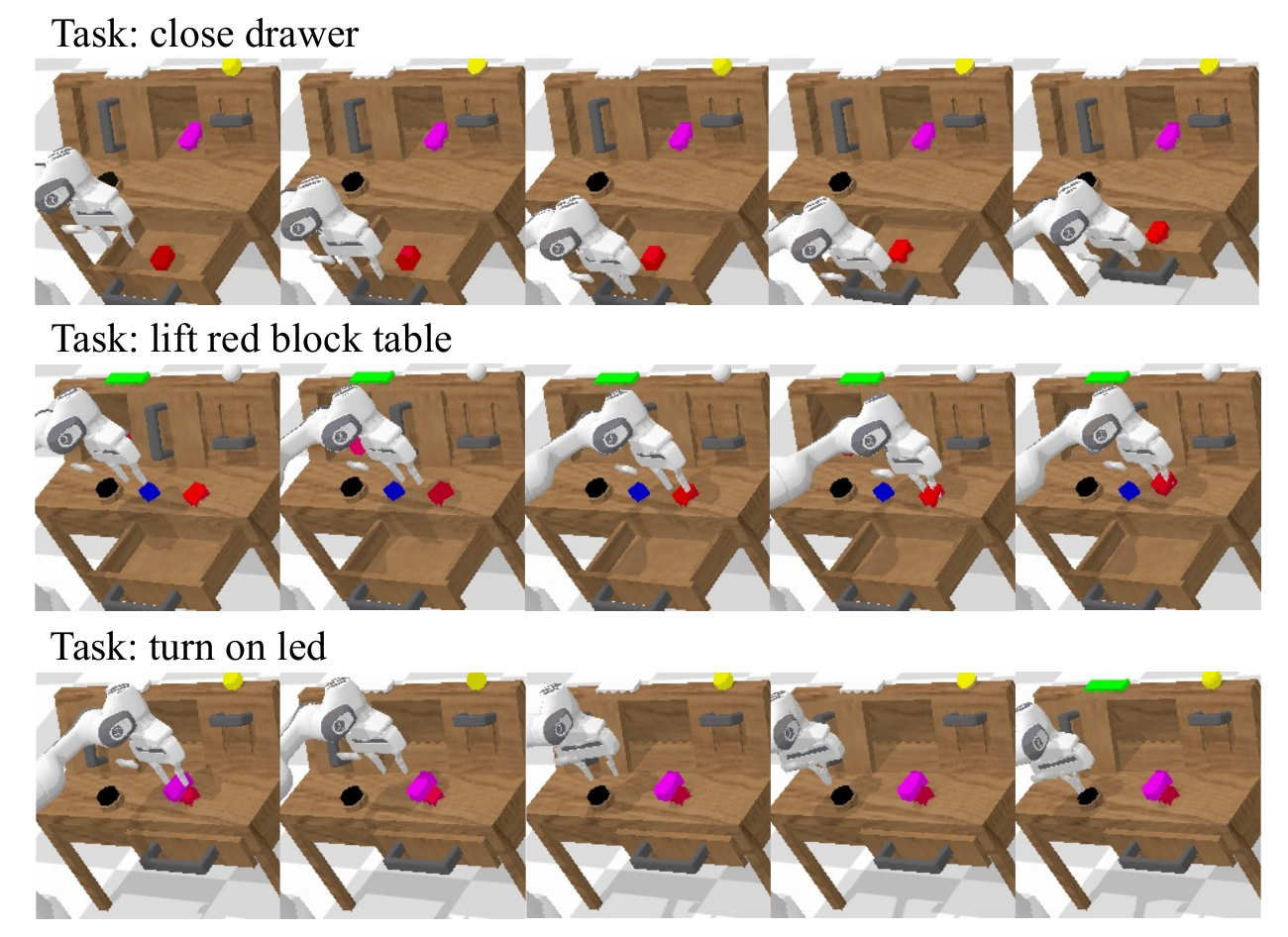}
    \caption{\textbf{Visualization of simulation evaluation tasks.} We visualize the simulation results of three different tasks, which demonstrates success of our model in multi-task learning.}
    \label{fig:exp_simulation}
\end{figure}

The results, as presented in Table 1, show that SSM-VLA achieves top performance on the ABC-D benchmark, outperforming a wide array of existing methods. Specifically, SSM-VLA surpasses direct prediction models (e.g., Roboflamingo~\cite{roboflamingo}, Dita~\cite{hou2025dita}), latent-to-real action models (e.g., Moto-GPT~\cite{chen2024moto}, UniVLA~\cite{bu2025learning}), and integrated visual foresight models (e.g., Seer~\cite{tianpredictive}, VPP~\cite{hu2024video}). We attribute this success to our model's unique cascaded architecture, which enables superior multi-task learning and generalization.
Figure~\ref{fig:exp_simulation} presents visualizations for three distinct scenarios. For each scenario, we show the results from five simulations, capturing a snapshot every five actions. Collectively, these results demonstrate our model's effectiveness in a multi-task learning setting.

\begin{figure}[h]
    \centering
    \includegraphics[width=0.95\columnwidth]{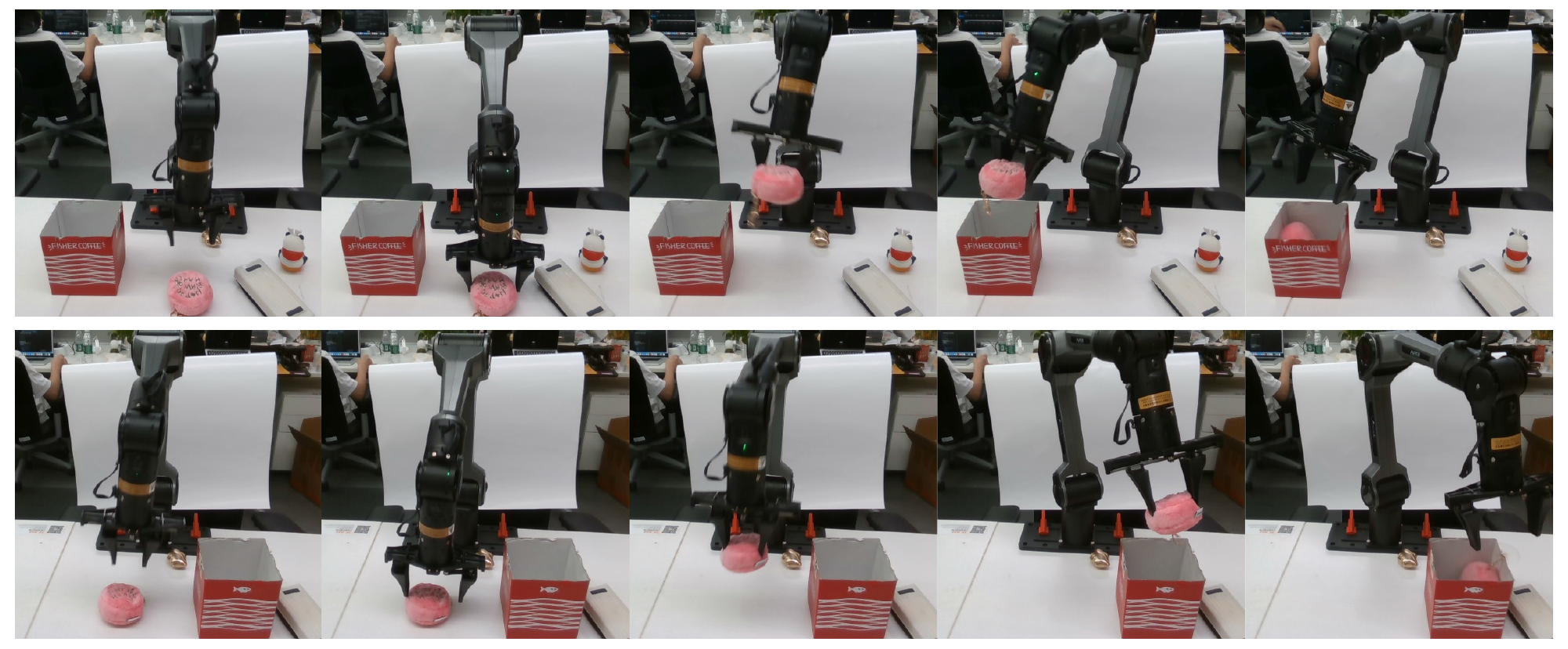}
    \caption{\textbf{Visualization of the real world experiments.} The model is asked to place the pink ball into the box. We show two samples with different layouts and chaos background. }
    \label{fig:exp_real}
\end{figure}

\subsection{Real World Experiments}
We evaluate our approach on a real-world robotic manipulation task using a single AgileX Piper robot, tasked with placing a pink toy into a box. The model is first pretrained on the large-scale Open-X-Embodiment dataset~\cite{o2024open,khazatsky2024droid,walke2023bridgedata}, and subsequently fine-tuned on 50 human-collected demonstrations. We masked the gripper camera input for real-world experiment because of limited condition. As demonstrated in Figure~\ref{fig:exp_real}, our method achieves successful deployment on the physical robot and exhibits strong generalization to real-world conditions, including cluttered and unstructured environments.

\subsection{Ablation Study}
\input{Tab/ablation_lam.tex}

To verify the benefit of our Farsighted-LAM, Multi-modal Synergistic Attention and Geometric Priors, we conduct an ablation study as shown in Table~\ref{tab:lam_ablation}. All ablation experiments are conducted on the CALVIN benchmark and run with the same training budget for consistency.
\subsubsection{Importance of Farsighted LAM Structure}
\begin{figure}[t]
    \centering
    \includegraphics[width=0.8\columnwidth]{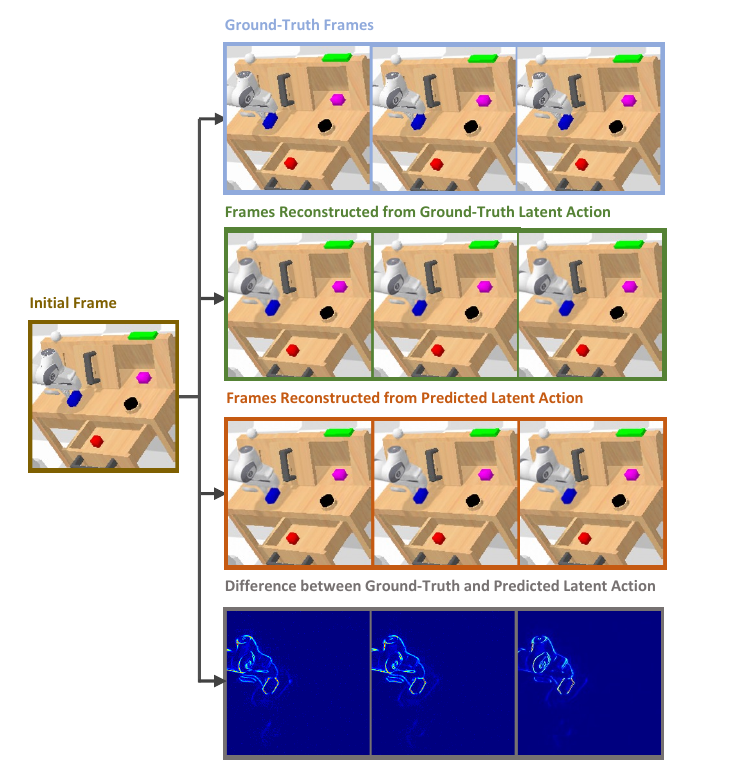}
    \caption{\textbf{Visualization of the latent action.} The similarity of these four rows demonstrates that 1) our Farsighted-LAM has learned the dynamic scene with spatial and motion awareness by ground-truth latent action $z_{t+k}$ (i.e., the second row) and 2) the latent action $\hat{z}_{t+k}$ (i.e., the third row) predicted by our VLA model aligns well with the ground-truth latent action.}
    \label{fig:exp_latent}
\end{figure} 
Our full model, which uses a 3-frame context (i.e., \textbf{Full Model}), achieves the best performance with an average task chain length of 4.38. In comparison, a more "direct" variant using only a single future frame (i.e., \textbf{LAM (1-frame)}) reduces the average length to 4.28, while a model where the LAM module is removed entirely (i.e., \textbf{w/o LAM}) causes a more significant performance drop to an average length of 4.23.
This two-tiered comparison indicates that: First, compared to predicting only a single future frame, using a 3-frame context introduces intermediate-state constraints and provides more continuous dynamic supervision, encouraging the LAM to learn a smoother and more physically consistent latent action representation; Second, removing the LAM entirely significantly reduces the average task chain length, showing that latent action modeling is a key component for long-horizon planning and successful execution.
Here we visualize the latent action by decoding it with Farsighted-LAM. As shown in ~\cref{fig:exp_latent}, given the initial frame $s_t^{rgb}$, the first row shows three ground-truth frames in the future. The frames shown in the second row are reconstructed with ground-truth latent actions. Given the pretrained Farsighted-LAM, it first apply the encoder to get the ground-truth latent action $z_t$ and then reconstructed the frame $\hat{s}_{t+k}$ with the decoder. The reconstructed frames look similar to the ground truth, precisely tracking the spatial and motion changes. This confirms that the Farsighted-LAM has learned the dynamic scene with the well-designed structure. In the third row, we reconstruct frames using the same decoder, but instead of ground-truth actions, we feed in the latent action predicted by the overall VLA model as described in ~\cref{subsec:latent_from_VLA}. Results show the second and third rows also look similar, which confirms that the latent action $\hat{z}_{t+k}$ predicted by our VLA model aligns well with the ground-truth latent action $z_{t+k}$. We visualize the difference between the second row and the third row in the fourth row.

\subsubsection{Effect of Multi-modal Synergistic Attention}
We compared the multi-modal synergistic attention mechanism with a baseline method that uses a simple token-level causal attention mask and both were trained with the same number of steps.
As shown in Table~\ref{tab:lam_ablation}, replacing our structured attention with this naive causal baseline (i.e., \textbf{Causal Atten.}) leads to a dramatic performance collapse, with the average sequence length falling from 4.38 to 3.70, which significantly demonstrates the importance of our synergistic design. The simple causal attention mechanism leads to information leakage across different modalities (vision, latent action, real action). In contrast, our structured mechanism allows only necessary components to attend to the information needed from their corresponding modalities, preventing them from "shortcut learning" from other modalities.

\subsubsection{Contribution of Geometric Priors}
\input{Tab/depth_details}
To investigate the contribution of explicit 3D geometric information, we train a policy variant without any depth supervision (i.e., \textbf{w/o Depth}) and perform an ablation study. Removing depth yields a consistent but moderate degradation across metrics, with the average task chain length decreasing slightly from 4.38 to 4.30.
To better characterize when depth helps, we further compare tasks with different degrees of geometric dependency in Table~\ref{tab:ablation_tasks_models}. On the more depth-critical \textit{push into drawer} task, removing depth supervision reduces the success rate from 79.1\% to 73.6\%. In contrast, on the largely color-driven \textit{push blue block right} task, the two methods perform similarly (i.e., 73.2\% vs. 73.9\%). These results suggest that explicit depth supervision mainly benefits tasks which require accurate 3D spatial reasoning, helping the policy better infer relative object poses and manipulation affordances.

%% file: Tab/calvin_main.tex
\begin{table*}[t]
    \vspace{1.2mm}
    \centering
    \small 
    \caption{Comparative evaluation on the CALVIN benchmark. Our method demonstrates state-of-the-art performance, surpassing all baselines with higher success rates for N-length task sequences and a greater average successful chain length.}
    \label{tab:main_calvin_results}
    \definecolor{highlightcolor}{gray}{0.9}
    
    \newcolumntype{C}{>{\centering\arraybackslash}X} 
    
    \begin{tabularx}{\textwidth}{l | *{5}{C} | C}
        \toprule
        \textbf{Method} & \multicolumn{5}{c|}{\textbf{Task completed in a row}} & \\
        \cmidrule(lr){2-6}
        & 1 & 2 & 3 & 4 & 5 & \textbf{Avg. Len.} $\uparrow$ \\
        \midrule
        Roboflamingo (ICLR 24)~\cite{roboflamingo}                    & 82.4 & 61.9 & 46.6 & 33.1 & 23.5 & 2.47 \\
        Susie (ICLR 24)~\cite{susie}                           & 87.0 & 69.0 & 49.0 & 38.0 & 26.0 & 2.69 \\
        Moto-GPT (ICCV 25)~\cite{chen2024moto}                        & 89.7 & 72.9 & 60.1 & 48.4 & 38.6 & 3.10 \\
        3D Diffusor Actor (CoRL 25)~\cite{ke20253d}               & 92.2 & 78.7 & 63.9 & 51.2 & 41.2 & 3.27 \\
        CLOVER (NeurIPS 24)~\cite{bu2024clover}                          & 96.0 & 83.5 & 70.8 & 57.5 & 45.4 & 3.53 \\
        Dita (ICCV 25)~\cite{hou2025dita} & 94.5 & 82.5 & 73.8 & 61.3 & 50.0 & 3.61\\
        RoboDual (CoRR 24)~\cite{bu2024robodual} & 94.4 & 82.7 & 72.1 & 62.4 & 54.4 & 3.66\\
        UniVLA (RSS 25)~\cite{bu2025learning}                          & 95.5 & 85.8 & 75.4 & 66.9 & 56.5 & 3.80 \\
        UP-VLA (ICML 25)~\cite{zhang2025up}                          & 92.8 & 86.5 & 81.5 & 76.9 & 69.9 & 4.08 \\
        Seer (ICLR 25)~\cite{tianpredictive}                            & 96.3 & 91.6 & 86.1 & 80.3 & 74.0 & 4.28 \\
        VPP (ICML 25)~\cite{hu2024video}                             & 95.7 & 91.2 & 86.3 & 81.0 & 75.0 & 4.29 \\
        \midrule
        \textbf{SSM-VLA}               & \textbf{97.6} & \textbf{94.1} & \textbf{88.3} & \textbf{81.8} & \textbf{75.9} & \textbf{4.38} \\
        \bottomrule
    \end{tabularx}
\end{table*}

%% file: Tab/ablation_lam.tex
\begin{table}[t]
    \centering
    \small
    \caption{Ablation study of proposed structure.}
    \label{tab:lam_ablation}
    \begin{tabular}{l|ccccc|c}
        \toprule
        \multirow{2}[2]{*}{Method} 
         & \multicolumn{5}{c|}{\textbf{Task completed in a row}} & \multirow{2}[2]{*}{\makecell{\textbf{Avg.}\\ \textbf{Len.}} $\uparrow$}\\
        \cmidrule(lr){2-6}
        & 1 & 2 & 3 & 4 & 5 &  \\
        \midrule
        \textbf{Full Model} & \textbf{97.6} & \textbf{94.1} & \textbf{88.3} & \textbf{81.8} & \textbf{75.9} & \textbf{4.38} \\
        \midrule
        LAM (1-frame) & 96.6 & 92.8 & 86.3 & 79.3 & 72.8 & 4.28 \\
        w/o LAM & 96.5 & 92.4 & 85.5 & 78.0 & 70.8 & 4.23 \\
        \midrule
        Causal Atten. & 93.8 & 83.7 & 73.3 & 63.4 & 55.6 & 3.70 \\
        \midrule
        w/o Depth & 97.3 & 91.3 & 86.0 & 80.7 & 74.4 & 4.30 \\
        \bottomrule
    \end{tabular}
\end{table}

%% file: Tab/depth_details.tex
\begin{table}[t]
    \vspace{1.2mm}
    \centering
    \small
    \caption{Comparison on depth-critical vs. non-depth-critical tasks.}
    \label{tab:ablation_tasks_models}
    \resizebox{\linewidth}{!}{%
    \begin{tabular}{l|ccc|ccc}
        \toprule
        \multirow{2}[2]{*}{Method} 
        & \multicolumn{3}{c|}{\textit{push into drawer}}
        & \multicolumn{3}{c}{\textit{push blue block right}} \\
        \cmidrule(lr){2-4}\cmidrule(lr){5-7}
        & \textbf{\makecell{Success\\Count}}
        & \textbf{\makecell{Total\\Count}}
        & \textbf{\makecell{Success\\Rate}}
        & \textbf{\makecell{Success\\Count}}
        & \textbf{\makecell{Total\\Count}}
        & \textbf{\makecell{Success\\Rate}} \\
        \midrule
        Full Model & 102 & 129 & 79.1\% & 52 & 71 & 73.2\% \\
        w/o Depth & 95 & 129 & 73.6\% & 51 & 69 & 73.9\% \\
        \bottomrule
    \end{tabular}%
    }
\end{table}

%% file: sec/5_conclusion.tex
\section{Limitations and Conclusion} 
Concurrent works~\cite{zhang2025dreamvla,yuan2025depthvla} have also highlighted the importance of depth priors. However, how to prevent overfitting during latent action model training remains a highly valuable research direction.
In this work, we introduced SSM-VLA, a novel architecture designed to address the critical limitations of existing Latent Action Models in capturing geometric and dynamic information. By synergistically combining geometrically aware DINOv2 visual features with multi-frame temporal modeling, our model better captures both static scene structure and motion dynamics. The Chain-of-Thought reasoning pipeline further improves prediction by explicitly modeling environmental changes before action selection. Our experimental results show that this structured prediction method significantly improves performance, sets a new benchmark in multiple VLA benchmark tests, and highlights the significant value of integrating spatiotemporal cognition into embodied intelligence systems.